\documentclass[conference]{IEEEtran}
\IEEEoverridecommandlockouts
\usepackage{cite}
\usepackage{amsmath,amssymb,amsfonts}
\usepackage{graphicx}
\usepackage{textcomp}
\usepackage{xcolor}
\usepackage{caption}
\usepackage{subcaption}
\usepackage{algorithm}
\usepackage{algpseudocode}
\usepackage{multirow}
\def\BibTeX{{\rm B\kern-.05em{\sc i\kern-.025em b}\kern-.08em
    T\kern-.1667em\lower.7ex\hbox{E}\kern-.125emX}}

\makeatletter 
\newcommand{\linebreakand}{%
  \end{@IEEEauthorhalign}
  \hfill\mbox{}\par
  \mbox{}\hfill\begin{@IEEEauthorhalign}
}
\makeatother 
    
\begin{document}

\title{Near Field iToF LIDAR Depth Improvement from Limited Number of Shots\\
}

\author{
\IEEEauthorblockN{1\textsuperscript{st} Mena Nagiub}
\IEEEauthorblockA{\textit{Dept. of Front Camera} \\
\textit{Valeo Schalter und Sensoren GmbH}\\
Bietigheim-Bissingen, Germany \\
mena.nagiub@valeo.com}
\and
\IEEEauthorblockN{2\textsuperscript{nd} Thorsten Beuth}
\IEEEauthorblockA{\textit{Dept. of Detection Systems} \\
\textit{Valeo Detection Systems GmbH}\\
Bietigheim-Bissingen, Germany \\
thorsten.beuth@valeo.com}
\and
\IEEEauthorblockN{3\textsuperscript{rd} Ganesh Sistu}
\IEEEauthorblockA{\textit{Dept. of Computer Vision} \\
\textit{Valeo Vision Systems}\\
Galway, Ireland \\
ganesh.sistu@valeo.com}
\and
\linebreakand
\IEEEauthorblockN{4\textsuperscript{th} Heinrich Gotzig}
\IEEEauthorblockA{\textit{Dept. Driving Assistance} \\
\textit{Valeo Schalter und Sensoren GmbH}\\
Bietigheim-Bissingen, Germany \\
heinrich.gotzig@valeo.com}
\and
\IEEEauthorblockN{5\textsuperscript{th} Ciarán Eising}
\IEEEauthorblockA{\textit{Dept. Of Electronic \& Computer Engineer} \\
\textit{University of Limerick}\\
Limerick, Ireland \\
ciaran.eising@ul.ie}
}

\maketitle

\begin{abstract} Indirect Time of Flight LiDARs can indirectly calculate the scene's depth from the phase shift angle between transmitted and received laser signals with amplitudes modulated at a predefined frequency. Unfortunately, this method generates ambiguity in calculated depth when the phase shift angle value exceeds $2\pi$. Current state-of-the-art methods use raw samples generated using two distinct modulation frequencies to overcome this ambiguity problem. However, this comes at the cost of increasing laser components' stress and raising their temperature, which reduces their lifetime and increases power consumption. In our work, we study two different methods to recover the entire depth range of the LiDAR using fewer raw data sample shots from a single modulation frequency with the support of sensor's gray scale output to reduce the laser components' stress and power consumption.
\end{abstract}


\section{Introduction}
Almost all major automotive manufacturers and relatively new players in the space, such as Google and Apple, are dedicating significant resources to the development of vehicle automation. For high levels of automation, the vehicles require a detailed 3D understanding of their environment. Conventional radar and ultrasonic have limited resolution (and range, in the case of ultrasound \cite{b15}). LiDAR uses LED light pulses coupled with accurate measurements of the reflection reception. Using this ``time of flight'' (ToF) distancing, the vehicle can extract the geometry of its surrounding.

Near Field LiDAR (NFL) sensors are currently used in many applications, including the automotive domain. In this domain, NFL is required to provide point clouds with a depth range of up 100 m and precision reaching up to millimeters. One category of these LiDARs is the indirect Time of Flight LiDARs (iToF). There are several classes of iTOF LiDARs, one being Amplitude Modulated Continuous Wave (AMCW), where the amplitude of the laser signal is modulated using a carrier modulation frequency (usually between 1 MHz and 24 MHz). Usually, the light source amplitude is modulated into a square waveform where $f$ is the modulation frequency, and $T$ is the modulation time. Then, the receiving sensor demodulates the received signal. The sensing elements are typically light-sensitive CMOS, so building enough charges from the reflected light signal requires some time, called the exposure time or integration time.

\begin{figure}[htbp]
\vspace{-0.3cm}
\centerline{\includegraphics[scale=0.35]{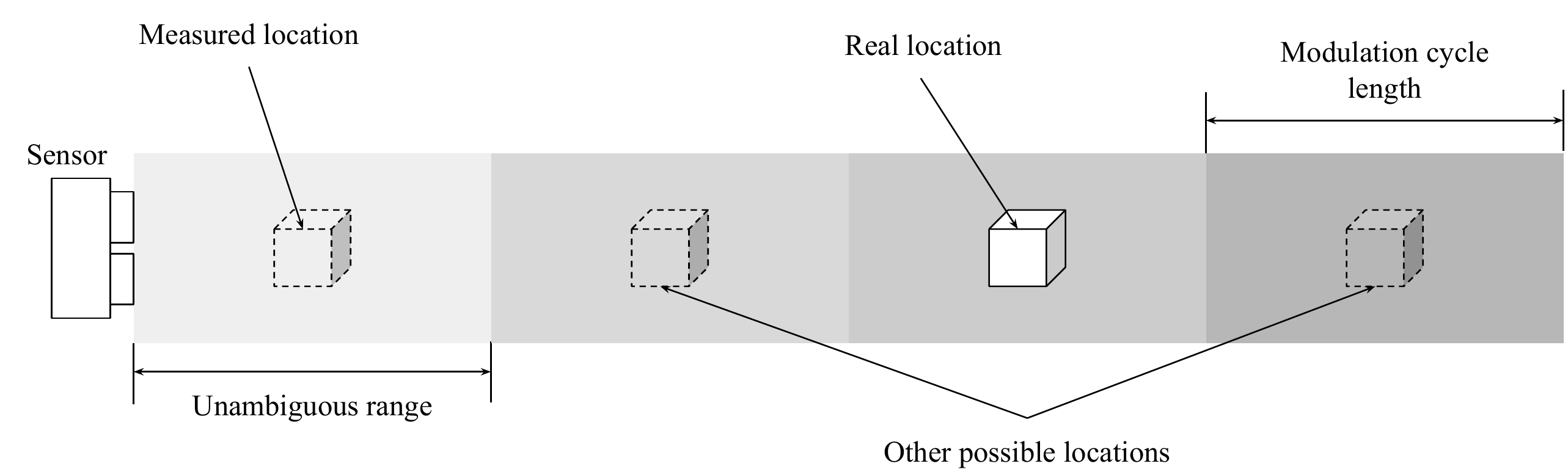}}
\vspace{-0.2cm}
\caption{Range ambiguity problem.}
\label{fig3}
\vspace{-0.2cm}
\end{figure}

AMCW iTOF Lidars indirectly calculate the distance of pixels through the phase shift angle between the transmitted and received signals. However, due to the modulation of the laser signal amplitude using periodic waves, it is possible to have phase wrapping of the phase shift angle when it exceeds $2\pi$. This phase wrapping leads to a problem called range ambiguity, as illustrated in figure \ref{fig3}.

The phase difference is calculated using cross-correlation between the received and transmitted signals, then transformed into the distance traveled. This method applies a feature extraction function to the received signal to search for the phase shift. The cross-correlation between the two signals generates a pattern function. Sampling the amplitude of the pattern function at equal steps over one period ($2\pi$), for example, $0^{\circ}$, $90^{\circ}$, $180^{\circ}$, and $270^{\circ}$, 
can give enough support points to calculate the phase shift difference angle ($\varphi$). 

Previous work has been done in this track to solve the problem of phase wrapping using computer vision and deep learning methods. Su et al. \cite{b3} have focused on solving the phase wrapping problem using deep convolutional layers directly from the raw samples. However, they used four raw data samples to predict the corresponding depth map. They have proposed an AutoEncoder architecture based on ResNet18. The model has been trained in a GANs configuration to solve the multi-path interference and ambiguity problems together. The solution has yet to consider reducing the number of shots required in each frame; they have also unwrapped the phase up to two cycles. Spoorthi et al. \cite{b1} have introduced a similar method to Su et al. to achieve phase unwrapping through deep learning but for Radar signals. Chen et al. \cite {b4} focused on solving the power consumption problem by reducing the exposure time. In their work Agresti et al. \cite{b5} focused on using deep learning to solve the problem of multi-path inference and have not considered the phase unwrapping but provided an interesting lightweight architecture that can be used to solve the phase wrapping problem as well.
Per our knowledge, previous research works focused on either solving the noise problems or unwrapping the phase shit issues up to a limited number of periods.

The contribution of this paper is based on the fact that iToF sensors can generate a gray scale image of the scene based on the ambient light. It is possible to predict a coarse depth map for the scene using convolutional neural networks from that gray scale image. In this case, we can unwrap the ambiguous range generated by the sensor using fewer raw data samples. This contribution provides several advantages:
\begin{itemize}
    \item It reduces the number of raw samples to reduce the laser components’ activation time and increases its lifetime.
    \item It reduces the consumed power required by the sensor to generate raw data samples.
    \item It enables the sensor to see beyond the expected range.
\end{itemize}

According to our previous study \cite{b6}, enriching the LIDAR point cloud using support from an additional sensor like a Camera is familiar. However, our significant contribution versus state-of-the-art is that we have used that method to reduce the number of LIDAR raw data samples required to build the point cloud.

\section{Proposed Method}
This research aims to create a method that utilizes computer vision to reduce the required laser shots to generate the final depth map. We will replace part of these shots with a gray scale image generated by the iTOF sensor imager based on the ambient light to achieve this goal. This feature is available in many iToF sensors. For example, Texas Instruments Sensor OPT8241 \cite{b8}, which is used by Su et al. \cite{b3}, can generate a 4-bit ambient image for the scene. In our study, we used our Valeo Near Field Lidar sensor \cite{b10}.
The sensor can generate a full-scale gray scale image for the ambient light environment, as illustrated in Figure \ref{Fig:epc660} (first row). In addition, a deep learning-based computer vision model is used to predict additional depth information, which is used to complement the missing details when the number of laser samples is reduced.

\subsection{Principle of Operation}
\begin{figure}[htbp]
     \vspace{-0.3cm}
     \centering
     \begin{subfigure}[b]{0.15\textwidth}
         \centering
         \includegraphics[width=\textwidth]{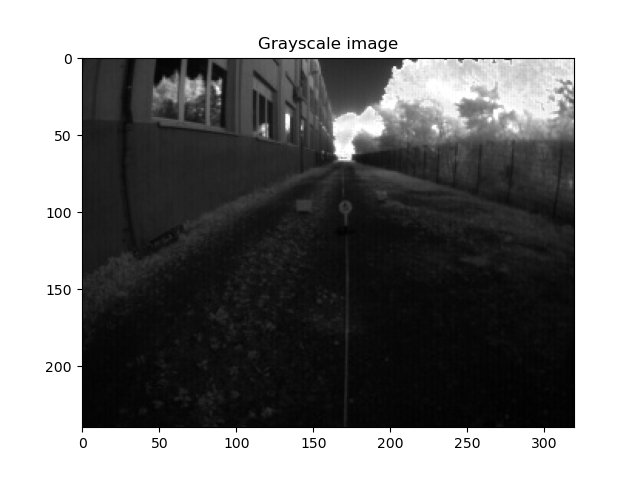}
     \end{subfigure}
     \hfill
     \begin{subfigure}[b]{0.15\textwidth}
         \centering
         \includegraphics[width=\textwidth]{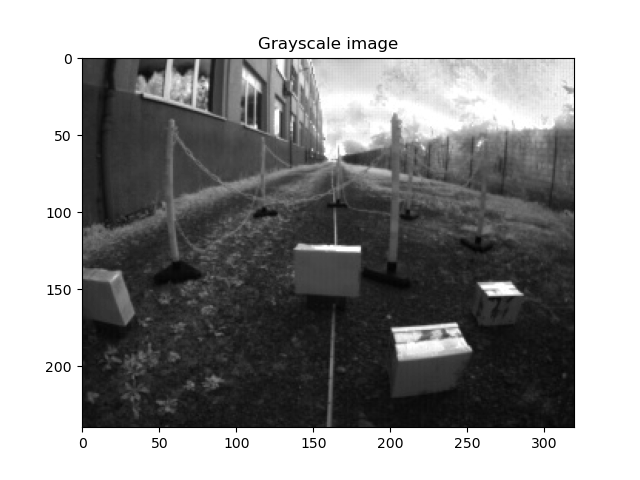}
     \end{subfigure}
     \hfill
     \begin{subfigure}[b]{0.15\textwidth}
         \centering
         \includegraphics[width=\textwidth]{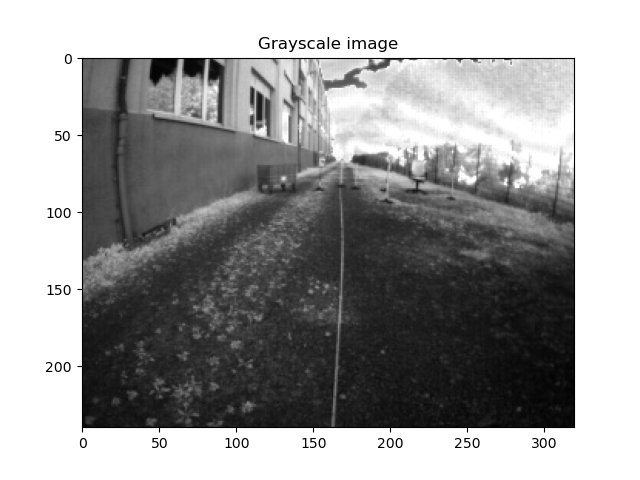}
     \end{subfigure}
         \hspace*{0.5mm}
     \begin{subfigure}[b]{0.15\textwidth}
         \centering
         \includegraphics[width=\textwidth]{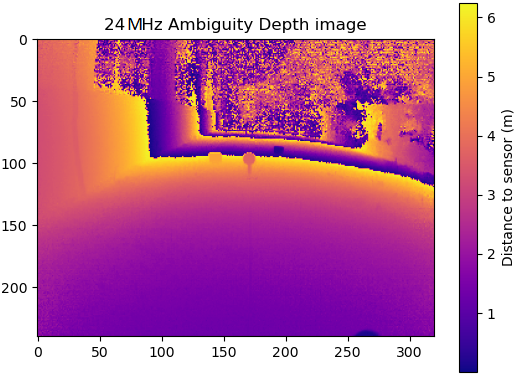}
     \end{subfigure}
         \hspace*{0.5mm}
     \begin{subfigure}[b]{0.15\textwidth}
         \centering
         \includegraphics[width=\textwidth]{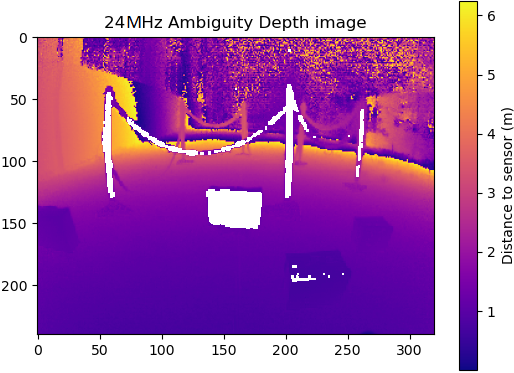}
     \end{subfigure}
     \hspace*{0.5mm}
     \begin{subfigure}[b]{0.15\textwidth}
         \centering
         \includegraphics[width=\textwidth]{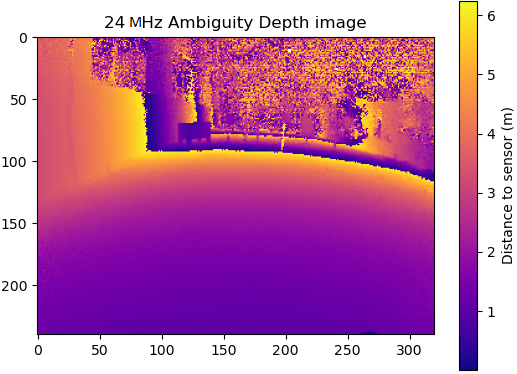}
     \end{subfigure}
     \caption{Examples of images captured by the Valeo NFL sensor (first row), and equivalent ambiguous depth maps captured at 24 MHz (second row). The maximum unambiguous range is 6.25 m. However, the sensor incorrectly detects objects beyond that range at distances between 0 and 6.25 m. Depth saturation is visible (white) in the center image.}
     \label{Fig:epc660}
\end{figure}

\begin{figure}[htbp]
\vspace{-0.5cm}
\centerline{\includegraphics[scale=0.32]{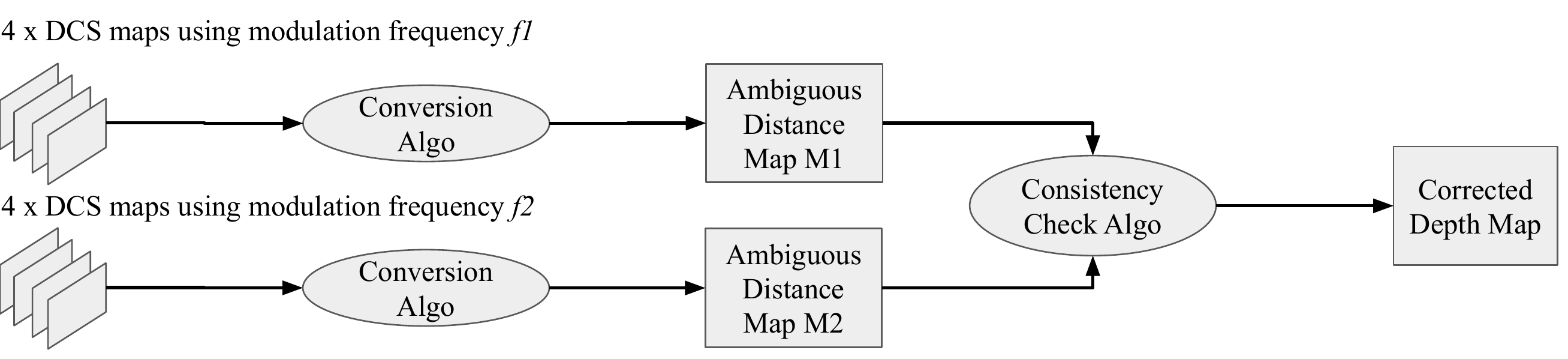}}
\caption{Depth map creation principle of operation.}
\label{fig5}
\vspace{-0.3cm}
\end{figure}

Figure \ref{fig5} shows how the sensor creates an accurate depth map from the raw data samples. The sensor captures four raw data samples (Differential Correlation Sample (DCS)) using modulation frequency $f_1$, typically 24 MHz. These four samples are combined to generate the first depth map $M_1$, with an ambiguity range $d_1$. The sensor then captures four additional raw data samples using modulation frequency $f_2$, usually lower than $f_1$, typically 10 MHz, and creates the second depth map $M_2$ with ambiguous depth range $d_2$. Since $f_1$ is higher than $f_2$, then $d_2$ is farther than $d_1$. Therefore, an ambiguous depth map $M_1$ is generated using high modulation frequency $f_1$, resulting in a high-accuracy depth map with a shorter unambiguous range. Figure \ref{Fig:epc660} (second row) shows examples of ambiguous depth maps. On the other hand, an ambiguous depth map $M_2$ is generated using low modulation frequency $f_2$, which results in a low-accuracy depth map but with a very long unambiguous range. Finally, the consistency check algorithm compares the two depth maps. Then, it generates the corrected depth map $M$ where the range ambiguity is removed up to 100 m according to the method described in Bulczak et al. \cite{b9}.

\begin{figure}[htbp]
\centerline{\includegraphics[scale=0.32]{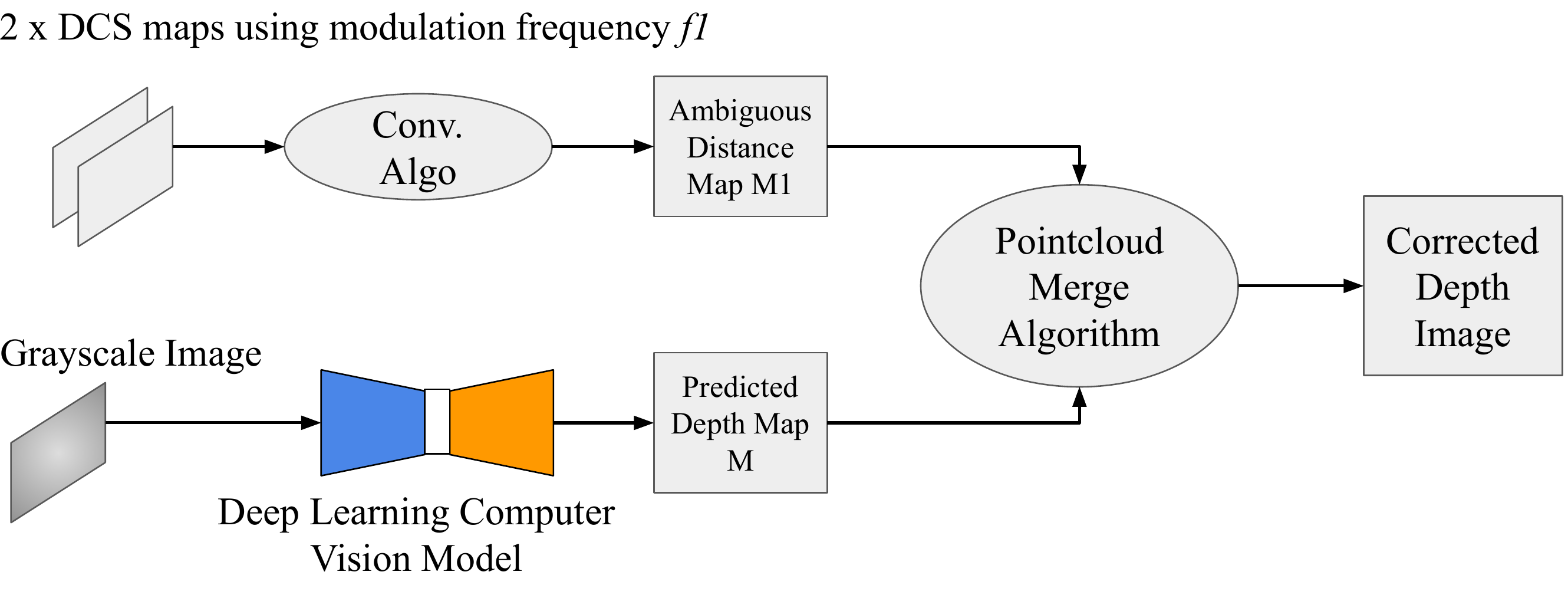}}
\caption{Our proposed method overview.}
\vspace{-0.3cm}
\label{fig7}
\end{figure}

To reduce the required number of laser shots, we propose a different method based on the gray scale image generated by the sensor for the ambient light, as illustrated in figure \ref{fig7}. The point merge algorithm implementation is dependent on the depth correction method whether using regression method or using segmentation method. In this case, we can replace the second group of 4 shots with the gray scale image, which does not require any laser activity. This replacement reduces the required raw samples from 8 shots to 4. We have also studied the possibility of reducing the first group of shots from 4 to 2 only to reduce the required shots even more. In this case, the phase shift angle is calculated as $\varphi = \tan ^{-1}\left ( \frac{S_2}{S_3} \right )$ .


\subsection{Method Architecture}
In our previous paper \cite{b6}, we have illustrated different architectures for depth prediction from monocular cameras or sensor constellations composed of a synchronized camera and laser sensor. Most of the proposed methods are based on AutoEncoder architecture following UNet architecture. In addition, some of these architectures have used LiDAR input to guide the final depth prediction. In this paper, we have studied two different architectures of the common architectures proposed.

\subsubsection{TofRegNet depth regression model}
\begin{figure}[htbp]
\centerline{\includegraphics[scale=0.32]{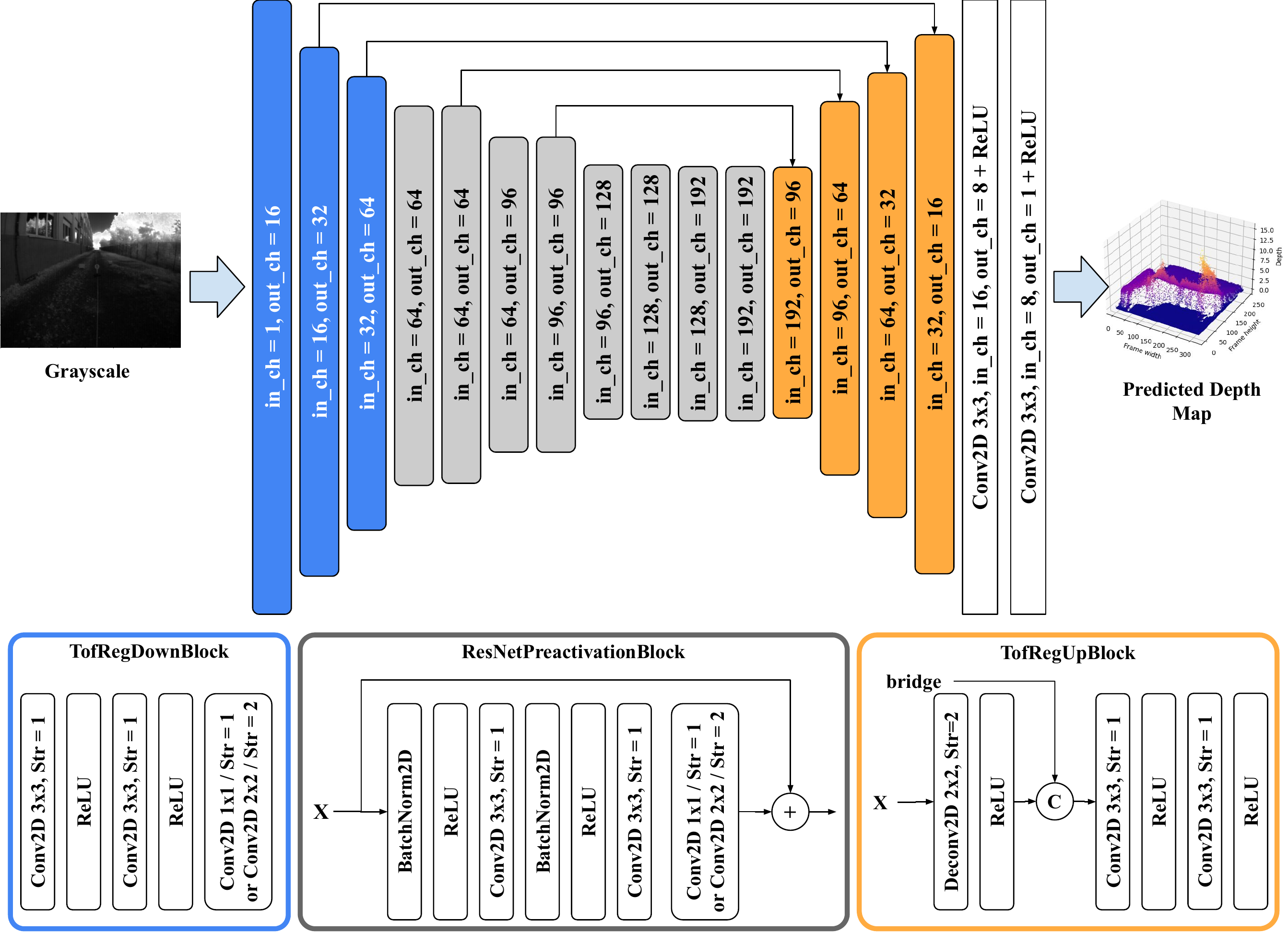}}
\caption{TofRegNet network architecture.}
\label{fig8}
\end{figure}

The first architecture is a supervised depth prediction network based on the AutoEncoder architecture. The depth prediction problem is solved as a regression problem, where the input to the model is the gray scale image generated by the sensor, and the final output is the predicted depth map. The ground truth for this model is the corrected depth image calculated by the sensor using the dual frequency method. The encoder is based on a feature extraction block followed by ResNet18 blocks and a decoder layer composed of 4 transposed convolution layers and four convolution layers in sequence. The final layer will predict the full-depth map. Figure \ref{fig8} explains the architecture of TofRegNet. 
The Pointcloud Merge Algorithm (Figure \ref{fig7}) is then used to correct the ambiguous depth map $M_1$ using 
(\ref{eq6}).
\begin{equation}
d_f^{i,j}=\left\{
	\begin{array}{ll}
		\left ( \left \lfloor d_p^{i,j}/d_u \right \rfloor \times d_u\right ) + d_{M1}^{i,j}, & d^{i,j}_{M1} < d_{sat} \\
		d^{i,j}_p, & d^{i,j}_{M1} \geq d_{sat}
	\end{array}
\right.
    \label{eq6}
\end{equation}
where $d_f^{i,j}\in D _{f}^{w\times h}$ is the final corrected depth pixel value at coordinates $(i, j)$, $d_p\in D _{p}^{w\times h}$ is the predicted depth pixel value at coordinates $(i, j)$, $d_{M1}\in D_{M1}^{w\times h}$ is the equivalent ambiguous depth pixel value in the ambiguous depth map $M_1$, $d_u$ is an unambiguous range of the depth map $M_1$ using modulation frequency $f_1$, $d_{sat}$ is the saturation range of the depth map $M_1$ using modulation frequency $f_1$ (normally defined by the sensor provider), $D _{f}^{w\times h}$ is the final corrected depth map of size $w\times h$ pixels, $D _{p}^{w\times h}$ is the predicted depth map by the model
, and $D _{M1}^{w\times h}$ is ambiguous depth map 
created using modulation frequency $f_1$. 

The loss function is designed to push the network to learn: 1) the depth prediction of the scene such that it would be similar to the ground truth, 2) the elimination of the pixels where depth saturation happens, and 3) the maximum depth range that the sensor can achieve. The first two points can be achieved if we penalize the model when the predicted image differs from the ground truth. For this reason, we have used two loss components. The first component is the scale invariant loss, and the second is the structural similarity index measurement loss \cite{b12}. The scale invariant loss is used to avoid the problem of depth saturation. Depth saturation could happen due to infinite depth regions, such as pixels representing the sky or open road, or retro-reflective surfaces reflecting a vast amount of light energy like metallic traffic signs. Scale invariant loss fixes this point through penalizing the model if it becomes dependent on the absolute depth of the points, and pushes the model to predict the pixel depth from the scene geometric characteristics in relation to surrounding points. For the third point, we have used a depth-guided loss function. In that loss function, we penalize the model more if there is a higher difference between the predicted depth and ground truth depth.
Nevertheless,  rather than doing this for all pixels in the image, we will perform this only for the $N$ pixels with the top depth value in the ground truth image. It will be possible to force the model to learn the maximum possible range of the ground truth depth map without affecting the other lower depth points using this depth-guided loss function. 
(\ref{eq7}) describes the details of the loss function.
\begin{equation}
    L_{Reg} = \alpha L_{SI} + \beta L_{SSIM} + \gamma L_{DG}
    \label{eq7}
\end{equation}
$L_{SI}$ is the scale invariant loss, given as
\begin{equation}
L_{SI} = \frac{1}{n}\sum_{i}^{n}d_{i}^{2}-\frac{1}{n^{2}}\left ( \sum_{i}^{n}d_{i} \right )^{2} \nonumber
\end{equation}
where $d\in D _{diff}^{w\times h}$, and $D _{diff}^{w\times h} = D _{GT}^{w\times h} - D _{P}^{w\times h}$; $D_{GT}^{w\times h}$ is the ground truth depth map, and $D_{P}^{w\times h}$ is the predicted depth map. $L_{DG}$ is the guided depth loss, given as
\footnotesize
\begin{equation}
L_{DG} = L_{2}\left ( argmax \left ( D _{GT}^{w\times h}, N \right ) - D _{P}^{w\times h} \left( argmax \left ( D _{GT}^{w\times h}, N \right ) \right )  \right )  \nonumber
\end{equation}
\normalsize
where $N$ is the number of pixels to be tested for depth guidance, is defined by a kernel of size $(w / 10, h / 10)$, and $w$ and $h$ are the width and height of the frame, respectively. The size of the kernel is selected empirically. $L_{SSIM}$ is the well-known structural similarity index measurement loss.  $\alpha$, $\beta$, and $\gamma$ are weights selected empirically to be 0.5, 0.4, and 0.1, respectively.

\subsubsection{TofSegNet depth segmentation model} \label{tofsegnet_sec}
The second architecture solves the problem as a segmentation problem rather than a regression. This architecture divides the total depth range into equally segmented spaces, such as depth bins. Each depth bin is assigned to a segmentation class, so the size of each depth bin is equivalent to the unambiguous range. The ground truth depth maps are segmented as well in the same way. Then rather than training the network to predict the final depth value, it will be trained to predict the segmentation class, which is equivalent to the depth bin. The predicted class value is multiplied by the unambiguous range to convert the depth bin back to a final depth value. Then the conversion algorithm is used to match the ambiguous depth map value to the depth class using (\ref{eq8})
\begin{equation}
d^{i,j}_f=\left\{
	\begin{array}{ll}
		\left ( d^{i,j}_{c}\times d_{u} \right ) + d^{i,j}_{M1}, & d^{i,j}_{M1} < d_{sat}\\
		\left ( d^{i,j}_{c}\times d_{u} \right ), & d^{i,j}_{M1} \geq d_{sat}\\
	\end{array}
\right.
    \label{eq8}
\end{equation}
given that $d^{i,j}_f\in D _{f}^{w\times h}$ as above, and $d^{i,j}_{M1}\in D_{M1}^{w\times h}$, and where $d^{i,j}_c\in D _{c}^{w\times h}$ is the predicted depth bin at coordinates $(i, j)$, $d_u$ is the unambiguous range according to frequency $f_1$, $D _{p}^{w\times h}$ is the predicted depth bins map by the model of size $w\times h$ pixels.

\begin{figure}[htbp]
\centerline{\includegraphics[scale=0.35]{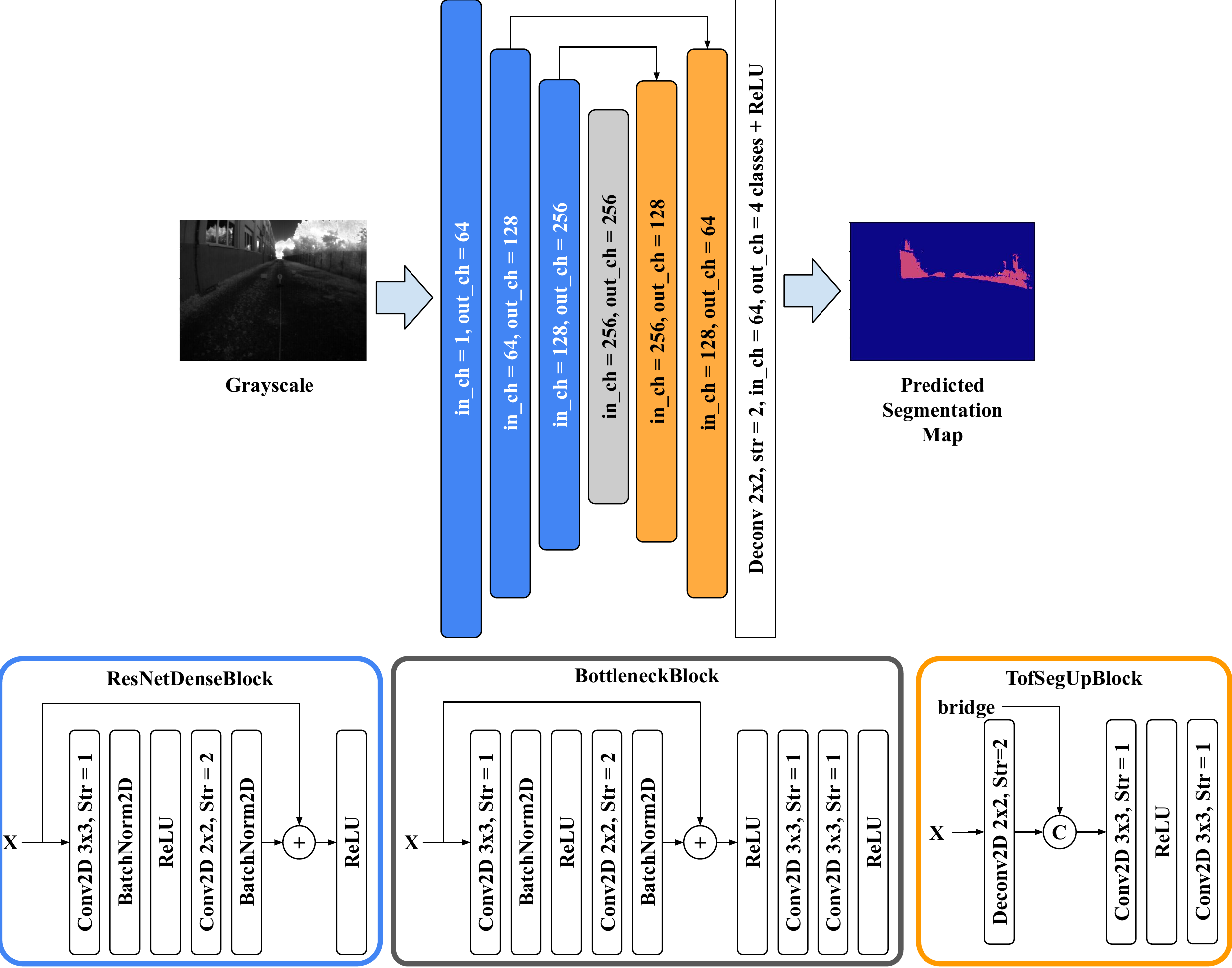}}
\caption{TofSegNet network architecture.}
\label{fig11}
\end{figure}

The semantic segmentation model based on auto encoder architecture structure is trained in a supervised fashion. The input to the model is the gray scale image generated by the sensor, and the final output is the predicted depth bin for each pixel. Ground truth for this model is the corrected depth image calculated by the sensor using the dual frequency method after being segmented to the correct depth bin. The encoder is based on ResNet18, followed by a decoder layer composed of 4 transposed convolution layers and four convolution layers in sequence. The final layer will predict the depth bins map; this model is called TofSegNet, and its architecture is described in Figure \ref{fig11}. 
The segmented depth map predicted by TofSegNet model is then used to correct the ambiguous depth map $M_1$ using the Segmented Point cloud Merge Algorithm (Figure \ref{fig7}).

The loss function is designed to push the network to learn the correct depth bins; the other two components defined in the previous section are not required since the segmentation process will eliminate the saturation problem implicitly. So, the loss function used here is the cross entropy loss function $L_{Seg} = L_{CE}\left ( D_{GT}^{w\times h}, D_{P}^{w\times h} \right )$. 

\section{Dataset}
\begin{figure}[htbp]
     \vspace{-0.3cm}
     \centering
     \begin{subfigure}[b]{0.15\textwidth}
         \centering
         \includegraphics[width=\textwidth]{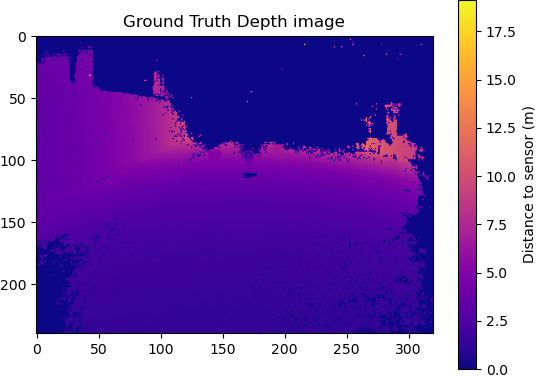}
     \end{subfigure}
     \hfill
     \begin{subfigure}[b]{0.15\textwidth}
         \centering
         \includegraphics[width=\textwidth]{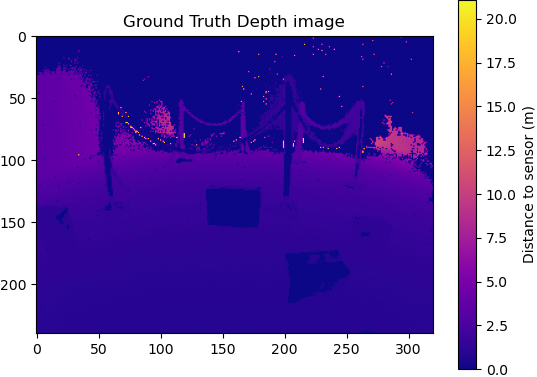}
     \end{subfigure}
     \hfill
     \begin{subfigure}[b]{0.15\textwidth}
         \centering
         \includegraphics[width=\textwidth]{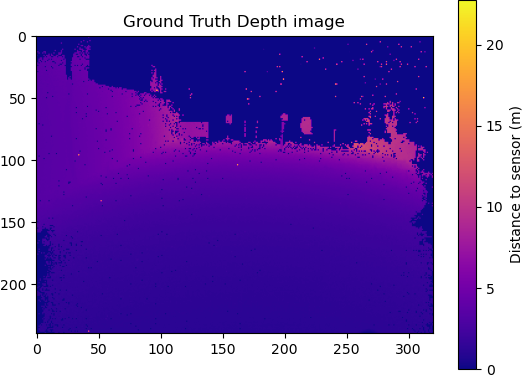}
     \end{subfigure}
        \caption{Sample of the dataset ground truth depth frames for the samples shown in Figure \ref{Fig:epc660}.}
        \label{fig18}
     \vspace{-0.3cm}
\end{figure}

The dataset used in this paper is generated using Valeo Near Field LiDAR sensors \cite{b10}. The dataset comprises 2048 frames created using different objects of different sizes that could be found in typical driving scenarios. 
Figure \ref{fig18} shows samples of the ground truth point cloud captured by the sensor using the dual frequency method. 

The dataset comprises 2048 samples, each consisting of a gray scale frame, a ground truth depth map generated using a dual frequency method, a depth map generated using 24 MHz frequency using 4 DCS, and a depth map generated using 24 MHz frequency using 2 DCS. The dataset is split into three parts, train, validation, and test using golden split values 80\% 10\% 10\%. The split is done using random selection. The dataset is augmented by flipping each frame around X-axis and Y-axis. In addition, the gray scale images are altered in contrast and brightness.

\section{Experiments and Results}
The experiments in this section aim to decide how to reduce the number of shots and which model is most suitable for shots reduction. In addition, we compare the number of required parameters of each model to find which model requires the least number of operations translated into lower power consumption. All models are trained on the same dataset, with frames of size 320 x 240. The implementation of the model is done using PyTorch running on Nvidia RTX 3080. The input gray scale image is 12-bit, while the ground truth depth map is 16-bit. Both are normalized into the value range [0, 1].

The TofRegNet model has been trained using an adaptive learning algorithm with the Adam optimizer, where the learning rate starts at 0.0001 and decays linearly down to 0.00009. Training is over 150 epochs with a batch size of 32. On the other hand, the TofSegNet model has been trained using a constant learning rate at 0.0001 with the Adam optimizer. The model has been trained over 140 epochs and batch size 64. 


In our previous paper \cite{b6}, we have explained that LiDAR-based methods use specific metrics according to the top papers based on famous automotive-based datasets like KITTI and Cityscape datasets. So, we have measured our models based on these metrics in our development system. The metrics measure the final result of the corrected depth map, so we can consider that all methods at the final stage are regression problems. Table \ref{tab3} summarizes the results of the experiments executed on the test dataset for the different models using four raw samples versus two raw samples. 

\begin{table*}[htbp]
\caption{Models experiments results and accuracy measurements}
\begin{center}
\begin{tabular}{|ll|c|c|c|c|c|c|c|c|}
\hline
\multicolumn{2}{|l|}{\multirow{3}{*}{\bf Experiment\textbackslash Metric}} & \multicolumn{4}{|c|}{\bf KITTI Depth Metrics \cite{b14}} & \multirow{3}{*}{\textbf{\begin{tabular}[c]{@{}c@{}}Prediction \\ Accuracy\\ (\%)\end{tabular}}} & \multirow{3}{*}{\textbf{\begin{tabular}[c]{@{}c@{}}Prediction \\ Precision\\ (\%)\end{tabular}}} & \multirow{3}{*}{\textbf{\begin{tabular}[c]{@{}c@{}}Correction \\ Accuracy\\ (\%)\end{tabular}}}  & \multirow{3}{*}{\textbf{\begin{tabular}[c]{@{}c@{}}Correction \\ Precision\\ (\%)\end{tabular}}}\\ \cline{3-6}
& & \textbf{\begin{tabular}[c]{@{}c@{}}RMSE\\ (Km)\end{tabular}} & \textbf{\begin{tabular}[c]{@{}c@{}}iRMSE\\ (1/Km)\end{tabular}} & \textbf{SqRel} & \textbf{AbsRel} &  & & &\\ \hline
\multicolumn{1}{|l|}{\multirow{2}{*}{\textbf{\begin{tabular}[c]{@{}l@{}}Using\\ 4 DCS\end{tabular}}}} 
& \textbf{TofRegNet}    & 2.02 & 4.3342 & 0.6032 & 0.6084 & 98.66\% & 99.27\% & 94.39\% & 99.06\% \\ 
\multicolumn{1}{|l|}{}  
& \textbf{TofSegNet} & 2.01 & 4.3134 & 0.6028 & 0.5969 & 99.56\% & 99.99\% & 94.44\% & 99.08\% \\ \hline
\multicolumn{1}{|l|}{\multirow{2}{*}{\textbf{\begin{tabular}[c]{@{}l@{}}Using\\ 2 DCS\end{tabular}}}} 
& \textbf{TofRegNet}    & 1.81 & 4.2654 & 0.4974 & 0.5692 & 98.66\% & 99.27\% & 94.93\% & 98.63\% \\ 
\multicolumn{1}{|l|}{}
& \textbf{TofSegNet}    & 1.78 & 4.2447 & 0.4878 & 0.553  & 99.61\% & 99.99\% & 95.05\% & 98.64\% \\ \hline 
\end{tabular}
\end{center}
\label{tab3}
\end{table*}

\begin{figure*}[htbp]
\begin{center}
\centerline{\includegraphics[scale=0.65]{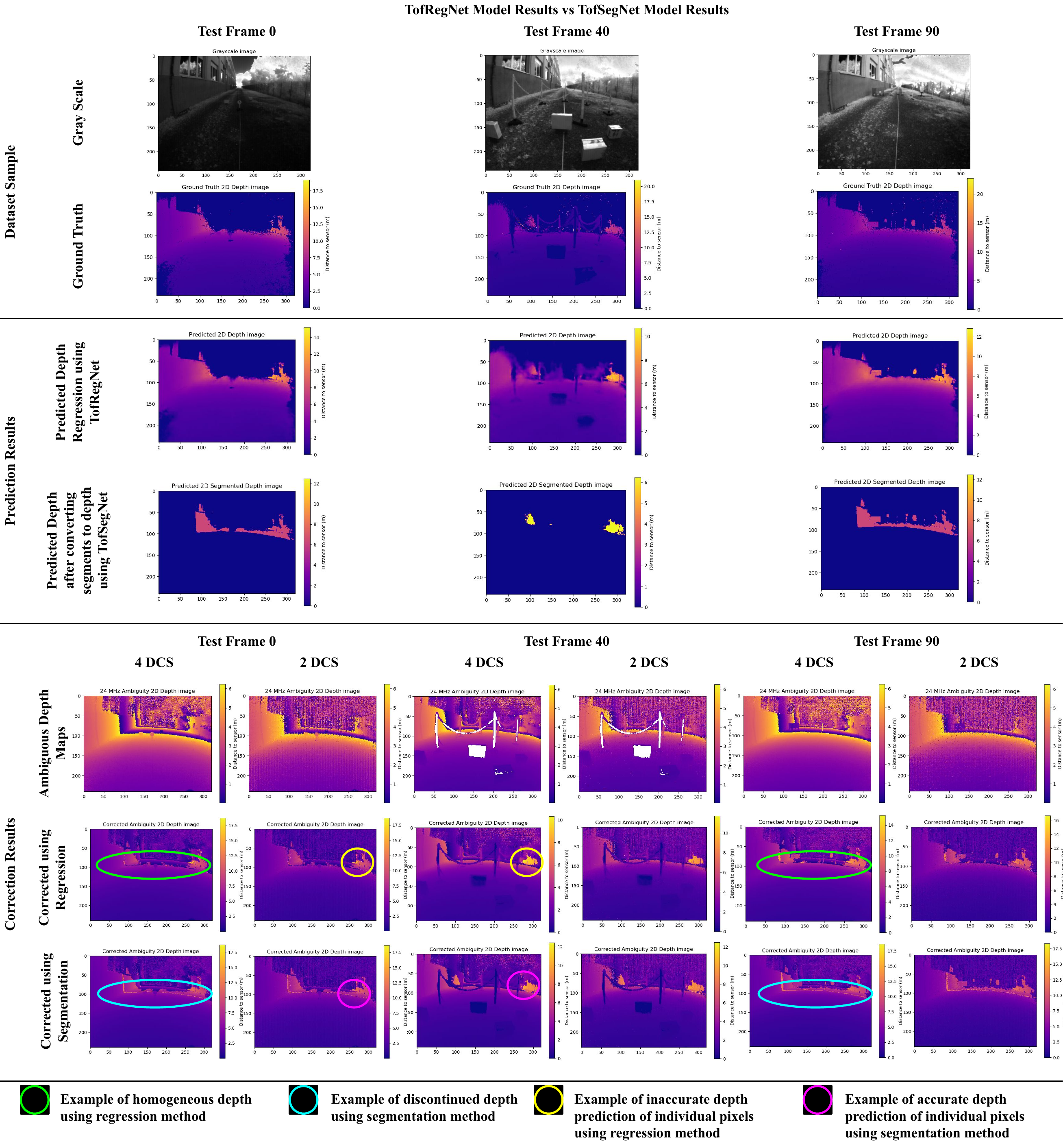}}
\caption{TofRegNet experiments results versus TofSegNet experiments results on 4 DCS versus 2 DCS.}
\label{fig20}
\end{center}
\end{figure*}

A detailed analysis of the results is provided in this section to draw visibility to the experiments performed. We compare the results of the experiments of each model when used to correct the ambiguity problem for ambiguous depth maps created using four raw samples versus maps created using only two raw samples. All the results are illustrated on test frames 0, 40, and 90 to simplify visually comparing the results. Figure \ref{fig20} shows the experiments executed using the TofRegNet model versus the TofSegNet model. The experiments are executed to correct ambiguous depth maps created using four raw samples versus maps created using only two raw samples. The figure is split into three sections. The first section has two rows; the first row shows the gray scale images of the three frames. Gray scale images are the same irrespective of the number of raw samples. The second row shows the ground truth depth generated by the sensor. Then the second section shows the prediction results of the TofRegNet model versus TofSegNet model. TofRegNet model predicts the depth as a regressed value, while TofSegNet model predicts the depth as a segmentation map. Then comes the exciting part, the third section, which shows how the predicted depth maps are used to correct the ambiguous depth maps. The first row shows the ambiguous depth maps to be corrected, which are generated using 4 DCS versus 2 DCS. These depth maps are created using the laser signal modulated at 24 MHz. A closer look at these frames clearly shows the problem of ambiguity where the depth of objects is wrapped every 6.25 m. Also, it shows that ambiguous depth maps created using 2 DCS suffer from a poorer quality and much noise compared to maps created using 4 DCS. The last two rows are the most interesting as they show the ambiguity problem correction using the depth regression versus depth segmentation. The depth of the objects is unwrapped to cover the whole depth range. As illustrated in the figure, corrected depth using depth regression maps is more homogeneous and continuous than segmentation maps. On the other hand the accuracy of the corrected individual vertices using segmented depth maps is higher than those corrected using regression depth maps. Also, comparing the correction results of the 2 DCS maps versus the 4 DCS maps, the results of the 4 DCS look less noisy than that using 2 DCS.

\section{Conclusion}
Based on the results of the two methods, it has been illustrated that the concept of point cloud correction can achieve promising performance. Comparing the two methods, it is clear that the TofRegNet model gives the best overall accuracy and precision and provides homogeneous depth maps. On the other hand, TofSegNet model provides accurate and precise individual vertices. In terms of the power reduction functionality, we can safely interpret the results as follows; it is possible to use only two raw data samples to build an accurate and precise depth map with support from the gray scale image. However, the two raw samples corrected depth map suffers from noise and inaccuracies due to the limited depth information compared to the four raw samples method. This noisy depth is due to the actual noise in the source depth map of the ambiguous map generated using the 24 MHz modulation frequency. This noise explains why the correction algorithm's performance in the two raw samples' depth maps is worse than the four raw samples' depth maps. Compared to the method developed by Su et al. \cite{b3}, our methods can predict corrected depth up to the second and third cycles of ambiguity, covering up to 17.5 m. In addition, we can correct the depth using only two raw samples.



\bibliographystyle{IEEEtran}
\bibliography{IEEEabrv,references}

\end{document}